# Use of 1D-CNN for input data size reduction of LSTM in Hourly Rainfall-Runoff modeling

Kei Ishida[1,2], Ali Ercan[3], Takeyoshi Nagasato[4], Masato Kiyama[5], Motoki Amagasaki[6]

1. International Research Organization for Advanced Science and Technology, Kumamoto University, 2-39-1 Kurokami, Kumamoto 860-8555, Japan
   Email: keiishida@kumamoto-u.ac.jp (Corresponding author)
2. Center for Water Cycle, Marine Environment, and Disaster Management, Kumamoto University, 2-39-1 Kurokami, Kumamoto 860-8555, Japan
3. Department of Civil and Environmental Engineering, University of California, Davis, One Shields Avenue, Davis, California 95616, USA.
   Email: aercan@ucdavis.edu
4. Graduated School of Science and Technology, Kumamoto University, 2-39-1 Kurokami, Kumamoto 860-8555, Japan
   Email: 217d8320@st.kumamoto-u.ac.jp
5. Faculty of Advanced Science and Technology, Kumamoto University, 2-39-1 Kurokami, Kumamoto 860-8555, Japan
   Email: masato@cs.kumamoto-u.ac.jp
6. Faculty of Advanced Science and Technology, Kumamoto University, 2-39-1 Kurokami, Kumamoto 860-8555, Japan
   Email: amagasaki@cs.kumamoto-u.ac.jp

**Abstract**

An architecture consisting of a serial coupling of the one-dimensional convolutional neural network (1D-CNN) and the long short-term memory (LSTM) network, which is referred as CNNsLSTM, was proposed for hourly-scale rainfall-runoff modeling in this study. In CNNsLTSM, the CNN component receives the hourly meteorological time series data for a long duration, and then the LSTM component receives the extracted features from 1D-CNN and the hourly meteorological time series data for a short-duration. As a case study, CNNsLSTM was implemented for hourly rainfall-runoff modeling at the Ishikari River watershed, Japan. The meteorological dataset, consists of precipitation, air temperature, evapotranspiration, and long- and short-wave radiation, were utilized as input, and the river flow was used as the target data. To evaluate the performance of proposed CNNsLSTM, results of CNNsLSTM were compared with those of 1D-CNN, LSTM only with hourly inputs (LSTMwHour), parallel architecture of 1D-CNN and LSTM (CNNpLSTM), and the LSTM architecture which uses both daily and hourly input data (LSTMwDpH). CNNsLSTM showed clear improvements on the estimation accuracy compared to the three conventional architectures (1D-CNN, LSTMwHour, and CNNpLSTM), and recently proposed LSTMwDpH. In comparison to observed flows, the median of the NSE values for the test period are 0.455-0.469 for 1D-CNN (based on NCHF=8, 16, and 32, the numbers of the channels of the feature map of the first layer of CNN), 0.639-0.656 for CNNpLSTM (based on NCHF=8, 16, and 32), 0.745 for LSTMwHour, 0.831 for LSTMwDpH, and 0.865-0.873 for CNNsLSTM (based on NCHF=8, 16, and 32). Furthermore, the proposed CNNsLSTM reduces the median RMSE of 1D-CNN by 50.2%-51.4%, CNNpLSTM by 37.4%-40.8%, LSTMwHour by 27.3%-29.5%, and LSTMwDpH by 10.6%-13.4%. Although the improvements in comparison to LSTMwDpH were relatively small for the low flow (< 25th %tile) and the middle flow (between 25the and 75th %tiles), the estimation of the high flow ($\geqq$ 75th %tile) and the peak flow ($\geqq$95th %tile) was clearly improved. The median of RMSE for the test period was reduced by 12.0%-14.7% for the high flow, and by 15.9%-18.5% for the peak flow. The proposed architecture CNNsLSTM would be an effective approach for flood management and hydraulic structure design mainly under climate change which requires river flow to be estimated from meteorological datasets.

**Highlights:**

- An architecture consisting of a serial coupling of CNN and LSTM was proposed.
- The CNN outputs and hourly inputs for a short duration were given to LSTM.
- The proposed architecture was applied to hourly rainfall-runoff modeling.
- Proposed approach improved the estimation accuracy, especially high and peak flows.

## 1. Introduction

Being one of the interconnected components of water cycle, runoff is the excess precipitation that flows over the Earth's surface into the rivers, lakes, and oceans. Runoff estimation is vital for management of water resources, flood preparedness, and aquatic environment. Runoff transports sediment, nutrients, chemicals, and pesticides; therefore, affects erosion and deposition processes, as well as, ecological and biological processes. Based on the modeling structure, rainfall runoff modeling can be characterized by a) empirical, or data driven, black box models; b) conceptual models; and c) physically based models (Devia et al., 2015). It can also be classified as lumped and distributed models with respect to its parameters as a function of space and time and deterministic and stochastic models based on the stochasticity within the system (Devia et al., 2015). Regressions, fuzzy-based networks, and artificial neural networks (ANNs) are initial examples of the data driven approaches (Kratzert et al., 2018).

Being a hot topic in various fields, deep learning has already been applied to address various issues in hydrology including rainfall-runoff modeling as shown by some review papers (Shen, 2018; Xu and Liang, 2021). LSTM is a type of recurrent neural network (RNN) developed by Hochreiter and Schmidhuber (1997) and Gers et al. (2000). RNN is a type of deep learning architecture that is suitable to time series modeling because RNN generates outputs in consideration of the order of the sequential input data. LSTM is an RNN architecture that has memory and gates, which enable to learn long-term dependencies between the sequential input data and the target data. Therefore, LSTM is suitable to time series modeling that has long-term dependencies. Some recent studies (Li et al., 2021; Liu et al., 2020; Tian et al., 2018) also utilized LSTM including its sequence-to-sequence form for flow forecasting with precipitation as input. In addition, Van et al. (2020) employed convolutional neural network (CNN) for rainfall-runoff modeling. CNN is another type of deep learning architecture that uses the convolution operation, which extracts features from the input in consideration of the connections among neighboring data in the input. They utilized the one-dimensional version of CNN (1D-CNN) to extract features along the temporal order in the input data. In addition, Kratzert et al. (2018) compared the model performance between LSTM and a well-known conceptual model, Sacramento Soil Moisture Accounting Model, which is coupled with the Snow-17 snow routine model. Their results showed that LSTM is comparable or has slightly higher accuracy compared to a conceptual model. Thus, deep learning has high potential for rainfall-runoff modeling. Therefore, it would be beneficial to develop a more accurate and efficient rainfall-runoff model with deep learning approaches.

Because the previous studies, discussed above, implemented a rainfall-runoff model by a single use of CNN or LSTM, this study aims to improve the estimation accuracy of rainfall-runoff modeling by developing a new architecture consisting of both CNN and LSTM. Daily- or monthly-scale rainfall-runoff modeling is needed for water resources management but hourly-scale modeling is necessary for flood management. As such, to be able to address both water resources and flood management problems, this study targets rainfall-runoff modeling at the hourly time scale over a watershed where there are long-duration dependencies between the meteorological variables, such as precipitation and air temperature, and the river flow. Meanwhile, most mountainous watersheds in middle or high latitude have snowfall during winter. The snow accumulation and melting processes cause long-duration dependencies between the meteorological variables and the river flow. It may be valuable to improve the estimation accuracy of hourly-scale rainfall-runoff modeling at such watersheds.

There are already many studies that couples CNN and RNN. Most architectures of the combined use of CNN and RNN may be able to be categorized into three groups. The first group uses the convolution operation within RNN. This type of architecture was originally developed with LSTM as RNN by Shi et al. (2015), which is known as ConvLSTM. The second group uses CNN and RNN in parallel and the inputs are given to CNN and RNN. Then, the outputs from CNN and RNN are combined, flattened, and given to a fully-connected layer (e.g. Li et al., 2019; Vidal and Kristjanpoller, 2020; Xu et al., 2020; Yu et al., 2021; Zang et al., 2020). The third group uses a serial coupling between CNN and RNN. For this type of architecture, CNN and RNN are frequently used for different dimensions. CNN is utilized to extract spatial features or features among multiple variables from the input data. RNN is utilized to process the given information along time (e.g. Chen et al., 2019; Kim and Cho, 2019; Le et al., 2019; S. Li et al., 2020; Sun et al., 2019; Yang et al., 2020). CNN and RNN can also be utilized in the same direction. Swapna et al. (2018) and Liao et al. (2021) gave one-dimensional input data sequence to CNN. Then, they passed the extracted features by CNN directly to RNN. Swapna et al. (2018) employed this approach to differentiate diabetes and normal Heart Rate Variability and Liao et al. (2021) proposed it for automatic modulation classification applications in wireless spectrum monitoring. In this case, CNN works to condense the information in the input data sequence and reduce its data length.

On the other hand, Ishida et al. (2021) improved the estimation accuracy of hourly-scale rainfall-runoff modeling with LSTM at a snow-dominated watershed by a multi-time-scale input approach. In general, the same temporal resolution is utilized between the input and the target data for LSTM. Contrarily, the proposed approach in this article parallelly gives the daily dataset for a longer duration together with hourly dataset for a shorter duration to LSTM. The daily data were utilized to obtain the long-duration dependencies between the input and the targeted river flow. The hourly data were utilized to obtain hourly scale fluctuations of the input variables. The approach can reduce the input data length of LSTM. For example, the general approach requires 8760 (hours) of the hourly data as input in order to deal with one-year duration dependencies between the input and target variables. Contrarily, the proposed approach is able to keep the input data length less than 365 (days) because the daily time series is used to deal with a long-duration dependencies and a long length of hourly data is not required as input. The proposed approach was originally developed to reduce the computation requirements by reducing the input data length of LSTM. However, the reduction in the input data length also improved the estimation accuracy of river flow at the hourly scale compared to LSTM only with hourly data as input.

Note that the previously proposed approach was successful, but it uses only daily data for a long duration. It means that the approach loses the hourly behaviors in the input time series for a long duration. It may be possible to improve the estimation accuracy of rainfall-runoff modeling if some data containing hourly behavior in the input time series are given to LSTM instead of the daily data. For this reason, this study proposes a similar architecture to the third group of CNN+LSTM architecture for the same dimension (G, Swapna et al., 2018; Liao et al., 2021) because it condenses the information in the input data sequence (i.e., distillation) and reduces its data length (i.e., dimensionality reduction). The hourly data for a long duration are given to 1D-CNN. Then, the extracted features by 1D-CNN are hourly time series data for a short-duration are given to LSTM together. In other words, the extracted features from 1D-CNN with a shorter data length are utilized instead of the daily input data in the previously proposed approach by Ishida et al. (2021).

The proposed architecture is implemented for hourly-scale rainfall-runoff model at a snow-dominated Ishikari River watershed (IRW), where there are long-duration dependencies between the meteorological variables and the river flow. This study utilizes precipitation, air temperature, evapotranspiration, and long- and short-wave radiation as the meteorological input datasets and the river flow at a gauging station near the outlet of the study watershed as the target dataset. Finally, the performances of the previous architectures are compared with the proposed architecture by comparing their computational accuracy.

## 2. Methodology

This study proposes an architecture consisting of a serial coupling of 1D-CNN and LSTM. 1D-CNN is utilized to distillate the features from the hourly times series and reduce data length. Then the extracted features from 1D-CNN are given to LSTM as input together with the hourly data for a short duration. This proposed architecture is called CNNsLSTM, where "s" represents a serial connection. For comparison and as a benchmark, the single use of 1D-CNN and LSTM are also utilized for the hourly-scale rainfall-runoff modeling. Furthermore, two types of input data are utilized for LSTM. First one uses only hourly data as input to LSTM, which is referred to be as LSTMwHour in this study. For LSTMwHour, the temporal resolution of the input dataset is the same as that of the target dataset, which is the traditional way of using LSTM. Secondly, daily and hourly datasets are parallelly given to LSTM, which was proposed by Ishida et al. (2021), and referred as LSTMwDpH herein. Lastly, the performance of parallel architecture of CNN and LSTM (the second group), referred as CNNpLSTM, is also investigated in comparison to proposed CNNsLSTM approach.

### 2.1. One-dimensional convolutional neural network (1D-CNN)

CNN consists of series of layers such as convolutional, pooling, and fully-connected layers. The output of a layer is called the feature map for CNN. The feature map of a layer is used as the input to the next layer. The operation conducted in a convolutional layer is called the convolution operation. The convolution operation for 1D-CNN is

$$v_i^p = \sum_{n=1}^{N}\sum_{m=0}^{M-1} k_m^p u_{(i+m)}^n + b^p \quad (i = 1, 2, \cdots, I) \quad (1)$$

where $u$ is the input, $v$ is the feature map, $k$ is the kernel, and $b$ is the bias, which is the learnable parameter in the convolution operation. $N$ is the channel number of the input and the kernel with the size of $M$ works as a filter. The number of kernels can be multiple. $p$ indicates the kernel number, which is the same as the channel number of the feature map. The kernels are applied to the input with an arbitrary length of stride. The zero-padding is frequently applied to both sides of the input in order to avoid losing information of the edge of the input. Therefore, the size of the feature map $I$ depends on the width of the padding and the stride.

The operation of a pooling layer, which is called the pooling operation, also outputs a feature map $v$ by applying a filter (kernel) to the input $u$. The pooling operation for 1D-CNN is

$$v_i^p = F\left(u_{(i+0)}^n, u_{(i+1)}^n, \cdots, u_{(i+M-1)}^n\right) \quad (i = 1, 2, \cdots, I) \quad (2)$$

where $F$ is the pooling function. As the pooling function, the maximum or the average operations are frequently utilized. Because the pooling operation simply extracts the feature map by the pooling function which works as the filter (kernel), it has no learnable parameters. Meanwhile, the channel number $p$ of the feature map is equal to the channel number $n$ of the input. Contrarily, the kernels are applied to the input with an arbitrary length of stride, similar to the convolution operation. The padding can also be utilized for the pooling operation. The size of the feature map $I$ of the pooling operation also depends on the width of the padding and the stride.

In a fully-connected layer, all the inputs are rearranged as a single vector, and then the linear transformation is applied to the vector. The output of a fully-connected layer forms a vector or a scalar. In addition to these layers, activation function is also frequently used in a CNN structure. An activation function is utilized after the operation of a layer and applied to each element of the channels of the feature map. An activation function does not change neither the number of the channels nor the size of the feature map.

For rainfall-runoff modeling, to generate river flow at time $t$, meteorological data from $t - T + 1$ to $t$ are given to the first layer of 1D-CNN. $T$ is the size of the input of the first layer. The number of the channels of the input to the first layer is equal to the number of the variables of the meteorological data. The size of the input $T$ depends on how long duration dependencies need to be considered. For example, when previous 5040 hours (approximately seven months) of the meteorological data affects the flow, the size of the input $T$ is set to 5040. Then, the time series of the flow is obtained by gradually shifting the time window of the input meteorological data.

### 2.2. Long short-term memory (LSTM) network

LSTM has at least one recurrent block, which is called LSTM block. The LSTM block generally consists of the following equations:

$$\boldsymbol{g}_i(s) = \sigma(W_{ii}\boldsymbol{x}(s) + \boldsymbol{b}_{ii} + W_{hi}\boldsymbol{h}(s-1) + \boldsymbol{b}_{hi}), (3)$$

$$\boldsymbol{g}_f(s) = \sigma(W_{if}\boldsymbol{x}(s) + \boldsymbol{b}_{if} + W_{hf}\boldsymbol{h}(s-1) + \boldsymbol{b}_{hf}), (4)$$

$$\boldsymbol{g}_o(s) = \sigma(W_{io}\boldsymbol{x}(s) + \boldsymbol{b}_{io} + W_{ho}\boldsymbol{h}(s-1) + \boldsymbol{b}_{ho}), (5)$$

$$\boldsymbol{g}_c(s) = \tanh(W_{ic}\boldsymbol{x}(s) + \boldsymbol{b}_{ic} + W_{hc}\boldsymbol{h}(s-1) + \boldsymbol{b}_{hc}), (6)$$

$$\boldsymbol{c}(s) = \boldsymbol{g}_f(s)\otimes\boldsymbol{c}(s-1) + \boldsymbol{g}_i(s)\otimes\boldsymbol{g}_c(s), \quad (7)$$

$$\boldsymbol{h}(s) = \boldsymbol{g}_o(s)\otimes\tanh(\boldsymbol{c}(s)), \quad (8)$$

where $\boldsymbol{g}_i$, $\boldsymbol{g}_o$, $\boldsymbol{g}_f$, and $\boldsymbol{g}_c$ are called the input gate, the output gate, the forget gate, and the cell input, respectively. $\boldsymbol{x}$ is the input vector, and $\boldsymbol{h}$ is hidden state. $W$ and $\boldsymbol{b}$ are the weights and biases, respectively. The weights and biases are different in each equation for the input vector and hidden state, respectively. They are differentiated by the two subscripts. The weights and biases are the learnable parameters in the LSTM block. σ, tanh, and ⊗ are the sigmoid function, hyperbolic tangent function, and Hadamard product, respectively.

The hidden state $h(s)$ is the information out of the LSTM block at time $s$. The input vector $x(s)$ at time $s$ and the hidden state $h(s-1)$ from the previous time $(s-1)$ are given to the LSTM block. When $s=1$, $h(s-1)$ is set to zero. The cell state $c$ is the memory, which keeps information with the block. The forget gate $g_f$ adjusts how much of the memory is taken over from the previous time step. The input gate $g_i$ regulates the information given by the cell input $g_c$. The output gate $g_o$ regulates the information out of the block. When the length of the input time series data (IDL) is set to $T$, the LSTM block is recurrently used $T$ times. Finally, the hidden state $h$ is linearly transformed to the output $y(t)$.

For rainfall-runoff modeling, meteorological time series data from $s=t-T+1$ to $t$ are used as input to generate river flow at time $t$. The size of the input vector $x$ is equal to the number of the variables of the meteorological data ($N$). To obtain a time series of the output, LSTM is repeatedly run with the meteorological data with the width $T$ of the time window.

For LSTMwHour, the hourly meteorological data during the previous 5040 hours are given to obtain hourly river flow at a certain time. Then, LSTM is repeatedly run with 5040 hours of the time window to generate a time series of hourly flow. LSTMwDpH proposed by Ishida et al. (2021) is also utilized for comparison in this study. This previously proposed approach uses the daily data together with the hourly data as input for hourly rainfall-runoff modeling. Furthermore, it uses the same variables for the daily and hourly data, as such, the size of the input vector of the meteorological data is $N$. Figure 1a shows the input vector time series for the case when the length of the daily and hourly time series is the same. H and D indicate hourly and daily, and the units of time $s^H$ and $s^D$ are hour and day, respectively. The time of the river flow output is $t^H$, on the day $t^D$. $T^H$ and $T^D$ indicate the length of the hourly and daily input time series data, respectively. Thus, the daily and hourly time series data are given as input in parallel.

## 2.3. Serial coupling of CNN and LSTM (CNNsLSTM)

This study proposes a new architecture using 1D-CNN and LSTM for hourly-scale rainfall-runoff modeling. As shown in Figure 2a, the proposed architecture serially connects CNN and LSTM and is referred as CNNsLSTM. Hourly time series data for a long duration are given to the CNN component and are passed through the multiple layers of the CNN component, and the last layer generates a feature map $\hat{v}$ with the size $= I$ and with the number of channels $= P$. The length of the hourly data for a short duration is set to $I$. Then, the hourly data for the short duration and the last feature map from CNN were combined as shown in Figure 2a. The combined data are given to the LSTM component. The IDL of LSTM is equal to $I$, and the size of the input vector is equal to $(N+P)$ where $N$ is the number of input variables.

Similar to the traditional single use of 1D-CNN, for rainfall-runoff modeling, $T$ hours of the meteorological time series data until the time $t$ (from $t-T+1$ to $t$) are given to the CNN component to generate river flow at time $t$ as shown in Figure 1b. The number of the channels of the input to the first layer is equal to the number of the variables of the meteorological data ($N$). Then, as described above, the $I$ hours of the meteorological data (from $t-I+1$ to $t$) are given to the LSTM component together with the feature map of the last layer of the CNN component.

a) LSTMwDpH

$$\begin{array}{c} x(t-T+1) \quad x(t-T+2) \quad \cdots \quad x(t-1) \quad x(t) \\ \begin{bmatrix} x_1^H(t^H-T^H+1) & x_1^H(t^H-T^H+2) & \cdots & x_1^H(t^H-1) & x_1^H(t^H) \\ x_2^H(t^H-T^H+1) & x_2^H(t^H-T^H+2) & \cdots & x_2^H(t^H-1) & x_2^H(t^H) \\ \vdots & \vdots & \cdots & \vdots & \vdots \\ x_N^H(t^H-T^H+1) & x_N^H(t^H-T^H+2) & \cdots & x_N^H(t^H-1) & x_N^H(t^H) \\ x_1^D(t^D-T^D+1) & x_1^D(t^D-T^D+2) & \cdots & x_1^D(t^D-1) & x_1^D(t^D) \\ x_2^D(t^D-T^D+1) & x_2^D(t^D-T^D+2) & \cdots & x_2^D(t^D-1) & x_2^D(t^D) \\ \vdots & \vdots & \cdots & \vdots & \vdots \\ x_N^D(t^D-T^D+1) & x_N^D(t^D-T^D+2) & \cdots & x_N^D(t^D-1) & x_N^D(t^D) \end{bmatrix} \begin{array}{l} \text{Hourly Inputs} \\ \\ \text{Daily Inputs} \end{array} \end{array}$$

b) CNNsLSTM

$$\begin{array}{c} x(t-I+1) \quad x(t-I+2) \quad \cdots \quad x(t-1) \quad x(t) \\ \begin{bmatrix} x_1(t-I+1) & x_1(t-I+2) & \cdots & x_1(t-1) & x_1(t) \\ x_2(t-I+1) & x_2(t-I+2) & \cdots & x_2(t-1) & x_2(t) \\ \vdots & \vdots & \cdots & \vdots & \vdots \\ x_N(t-I+1) & x_N(t-I+2) & \cdots & x_N(t-1) & x_N(t) \\ \hat{v}_1^1 & \hat{v}_2^1 & \cdots & \hat{v}_{I-1}^1 & \hat{v}_I^1 \\ \hat{v}_1^2 & \hat{v}_2^2 & \cdots & \hat{v}_{I-1}^2 & \hat{v}_I^2 \\ \vdots & \vdots & \cdots & \vdots & \vdots \\ \hat{v}_1^P & \hat{v}_2^P & \cdots & \hat{v}_{I-1}^P & \hat{v}_I^P \end{bmatrix} \begin{array}{l} \text{Short Hourly Inputs} \\ \\ \text{Feature Map from CNN} \end{array} \end{array}$$

Figure 1. Input data structure for a) LSTMwDpH and for b) the LSTM component of CNNsLSTM

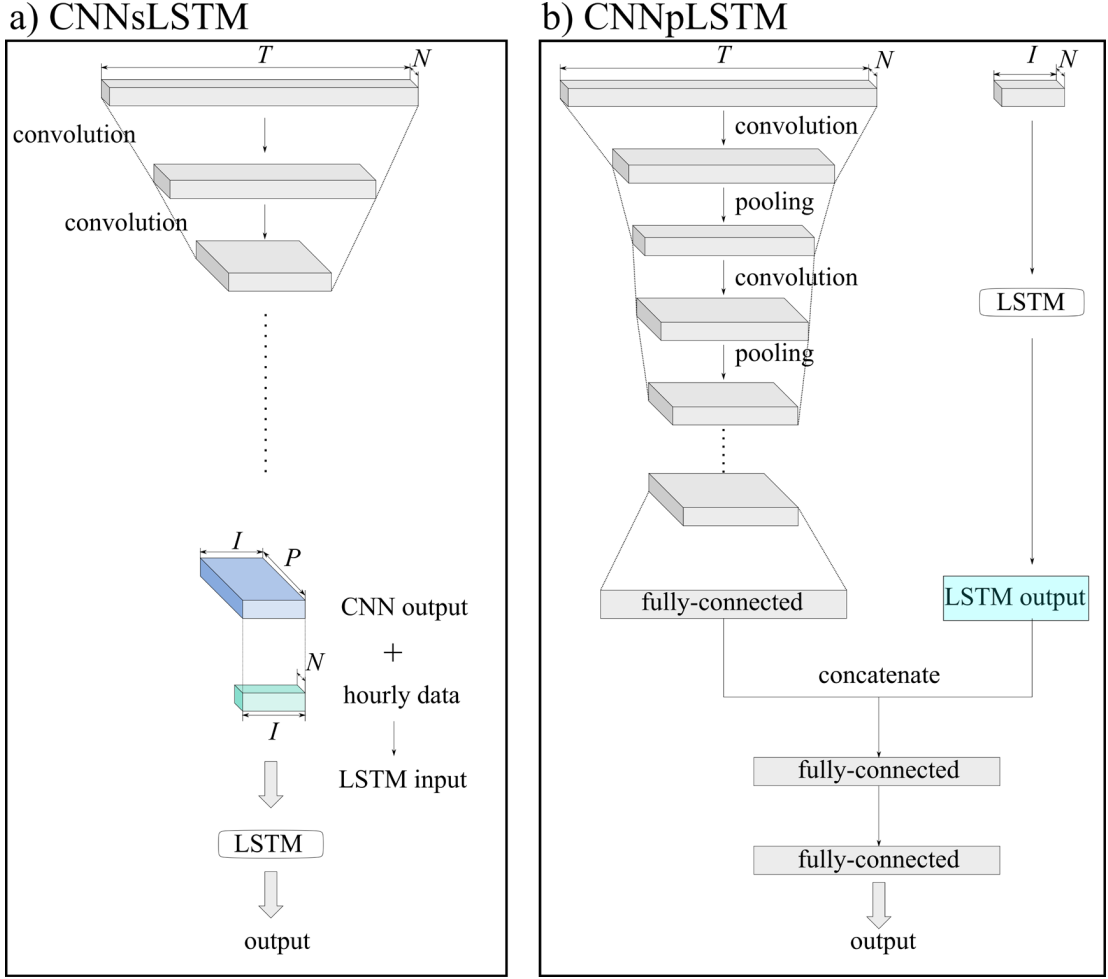

Figure 2. Architectures of CNNsLSTM and CNNpLSTM.

Notably, this study sets the size of hourly input time series data for a short duration, which is the same as the size of the feature map out from the CNN component. However, they are not necessary to be the same. If there is a difference between them, the zero-padding method can be utilized to align the length of hourly input data for a short duration same as that was done for LSTMwDpH in the previous study (Ishida et al., 2021).

### 2.4. Parallel use of CNN and LSTM (CNNpLSTM)

For comparison, this study also implements a rainfall-runoff model by a parallel architecture of 1D-CNN and LSTM, which is refereed to be as CNNpLSTM. As shown in Figure 2b, hourly input data for a long duration and for a short duration are given to the CNN and LSTM components, respectively. The outputs from the CNN and LSTM components are concatenated, and then given to a fully-connected layer. After passing several fully-connected layer, CNNpLSTM generates the output.

Unlike CNNsLSTM, the length of the hourly input time series data to the LSTM component of CNNpLSTM is flexible. For example, the hourly input data for the same duration can be given to the LSTM as well as CNN. However, for comparison, the lengths of the hourly input data to the CNN and LSTM components of CNNpLSTM, respectively, are set to the same as those of CNNsLSTM in this study.

### 3. Case Study

#### 3.1. Study Area and Data

The Ishikari River watershed (IRW) is located in the Hokkaido region, which is the northmost region in Japan (Figure 3). The center of IRW is at 43°26'06''N 142°05'06''E. Ranging from 0 m to 2,003 m elevation, IRW has 14,330 km² of the catchment size. Due to its latitude, it is cold enough to receive snowfall during winter to early spring (from October to March), especially in high elevations. As such, the flow at the Ishikari River generally increases from spring to early summer (from March to June) due to snow melt.

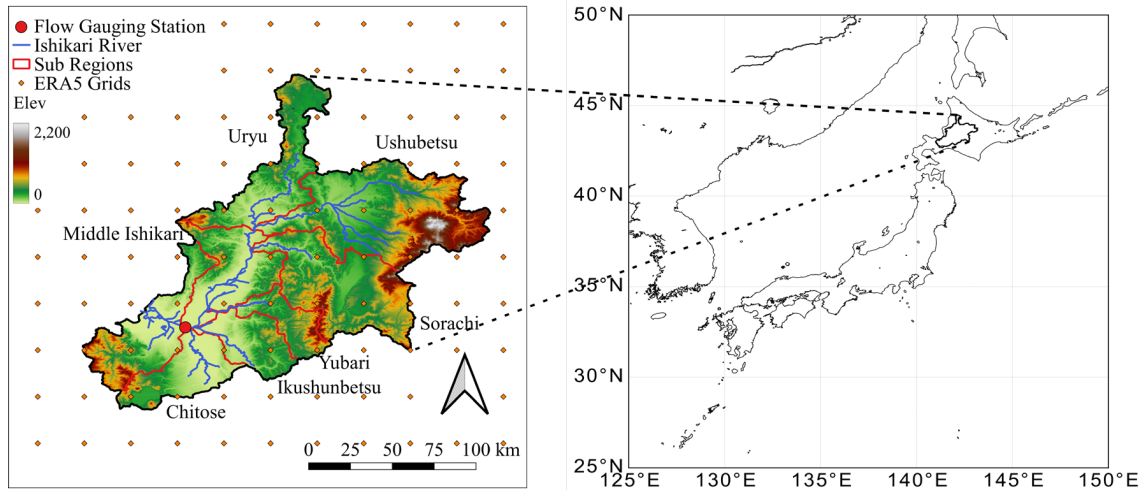

Figure 3. Study Area: The Ishikari River watershed

This study selected the hourly river flow data at the Ishikari Ohashi gauging station as the target data of the rainfall-runoff modeling. The Ishikari Ohashi gauging station is located at 26.60 km from the river mouth of the Ishikari River, as shown in Figure 3. The Ministry of Land, Infrastructure, Transport, and Tourism (MLIT) of Japan provides the hourly river flow data at the Ishikari Ohashi station from 1998 to 2019 on their Water Information System (MLIT, 2021)

Precipitation data were obtained from a gridded observed precipitation dataset, which is called the Radar Raingauge-Analyzed Precipitation (RRAP) generated by the Japan Meteorological Agency. This study obtained 1-km spatial resolution precipitation data from RRAP from 2006 to present. Hourly river flow will be affected by the spatial variability of precipitation within the watershed. Therefore, the study watershed upstream from the Ishikari Ohashi gauging station was separated into seven sub-regions as shown in Figure 3. Then, the area-averaged precipitation was calculated at each sub region from the 1-km resolution gridded data.

In the previous study (Ishida et al., 2021), only precipitation and air temperature were utilized as input. Contrarily, this study utilized not only air temperature and precipitation, but also evapotranspiration and long-and short-wave radiation as input. These datasets were obtained from the ERA5 reanalysis dataset (Copernicus Climate Change Service, 2017; Hersbach et al., 2020). The ERA5 reanalysis dataset is simulated data by three-dimensional atmospheric model that synthesized to various observation data. It contains not only meteorological data on surface as well as three-dimensional atmospheric data. The temporal resolution of ERA5 is hourly. The spatial resolution is 0.25º x 0.25º (approximately 20 km x 20 km) over the target watershed. The grids of ERA5 around IRW are shown in Figure 3. ERA5 is available from 1979 to present. The catchment area of the Ishikari Ohashi station is covered by 43 grid points. The area-averaged values of air temperature, evapotranspiration, and long- and short-wave radiation were calculated at the catchment area. It is noted that these variables were not obtained at the sub-regions unlike the precipitation. Air temperature has less spatial variability, and evapotranspiration and long- and short-wave radiation are less effective to river flow compared to precipitation. Because LSTM requires more computation resources with larger number of input datasets, the number of input datasets were restricted by the available computation resources.

### 3.2. Model Implementation

This study proposed an architecture CNNsLSTM. For comparison, four architectures, 1D-CNN, LSTMwHour, LSTMwDpH, and CNNpLSTM, are also implemented for hourly-scale rainfall-runoff modeling at the study watershed to model hourly river flow at the Ishikari Ohashi gauging station. The target period is set to 12 years from 2007 to 2019 based on the availability of the used input and target data as described above. This study utilized the area-averaged precipitation at the seven sub-regions and the area-averaged air temperature, evapotranspiration, and long- and short-wave radiation at the catchment of the flow gauging station as input. The area-averaged values of observed precipitation, air temperature, evapotranspiration, and short- and long-wave radiation at the catchment and observed river flow data at the Ishikari Ohashi station from 2007 to 2019 are illustrated in Figure 4. For all the architectures except LSTMwDpH, the length of the hourly input data length for a long duration is set to 5040 (hours), which is equal to 210 days and approximately seven months. Then, the length for a short duration is set to 210 (hours). Meanwhile, the lengths of the hourly and daily input time series for LSTMwDpH are set to 210 (hours) and 210 (days), respectively.

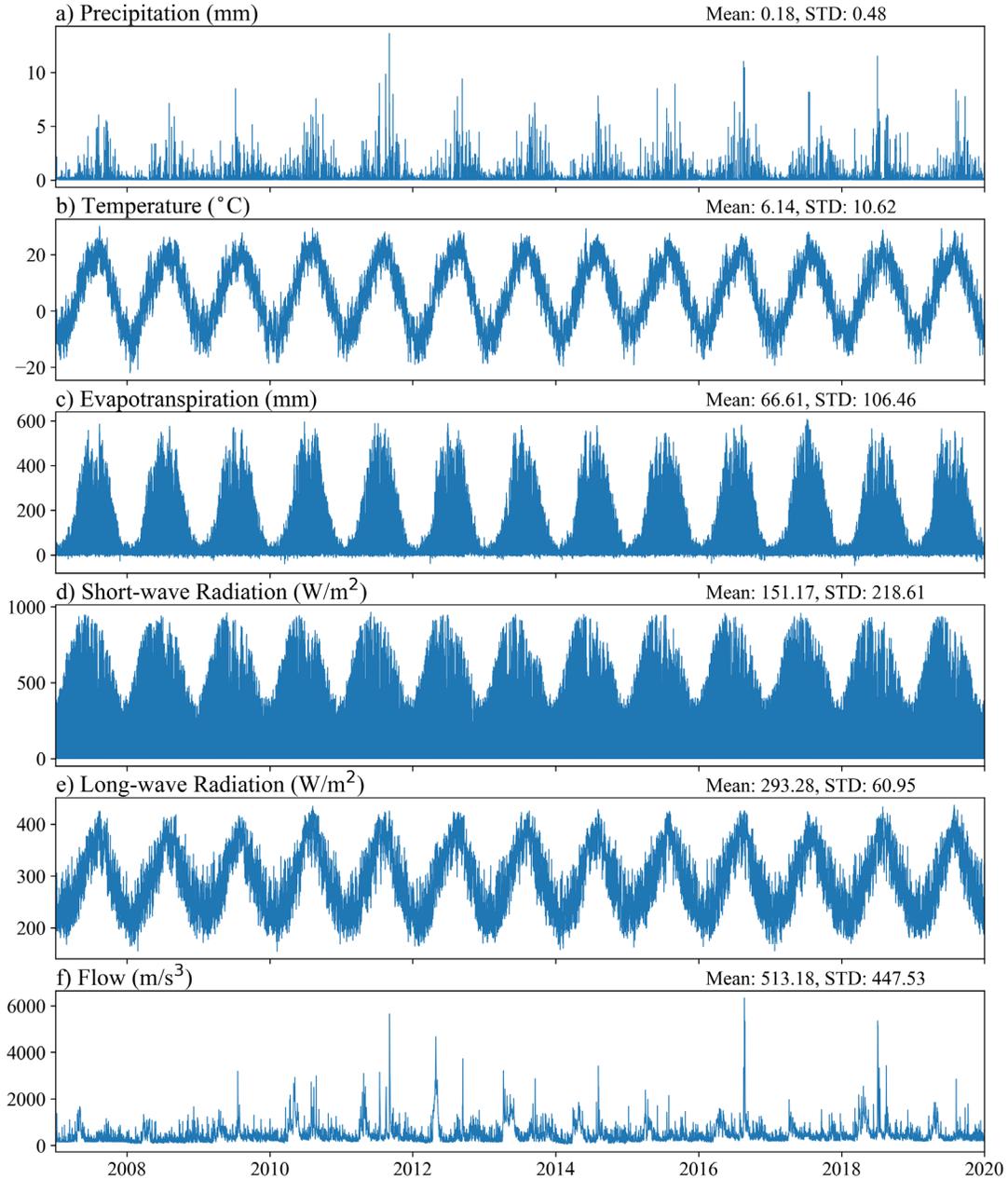

Figure 4. Area-averaged values of observed precipitation, air temperature, evapotranspiration, and short- and long-wave radiation over the catchment area and observed river flow data at the Ishikari Ohashi station.

Table 2 Configuration of the layers of 1D-CNN: The number of the channel of the feature map (NCH) or the number of the hidden nodes (NHN), the kernel size, the stride, and the padding width.

| Layer | NCH/NHN | Kernel | Stride | Padding |
|---|---|---|---|---|
| Conv1 | 8/16/32 | 6 | 3 | 3 |
| MaxPool1 | 8/16/32 | 4 | 2 | 2 |
| Conv2 | 16/32/64 | 4 | 2 | 2 |
| MaxPool2 | 8/16/32 | 4 | 2 | 2 |
| Conv3 | 32/64/128 | 4 | 4 | 2 |
| MaxPool3 | 32/64/128 | 4 | 2 | 2 |
| FC1 | 256 | - | - | - |
| FC2 | 128 | - | - | - |
| FC3 | 1 | - | - | - |

Table 1 Configuration of the layers of the CNN component of CNNsLSTM: The number of the channel of the feature map (NCH), the kernel size, the stride, and the padding width.

| Layer | NCH | Kernel | Stride | Padding |
|---|---|---|---|---|
| Conv 1 | 8/16/32 | 6 | 3 | 3 |
| Conv 2 | 16/32/64 | 4 | 2 | 2 |
| Conv 3 | 32/64/128 | 4 | 4 | 0 |

For the single use of 1D-CNN, three convolution layers, three max pooling layers, and three fully-connected layers are utilized. The configurations of the layers are tabulated in Table 1. After passing through the three sets of the convolution and max-pooling layers, the given data are passed to the first fully-connected layer. Three numbers of the channels of the feature map of the first layer (NCHF) were tried in this study: 8, 16, and 32. Then, the number of the channels of the feature map (output) is doubled at each convolution layer. Meanwhile, an activation function is applied after each convolution layer. As the activation function, the Rectified Linear Unit (ReLU) is employed.

For LSTMwHour and LSTMwDpH, the number of the layers and the length of the hidden and cell states is set to one and 30, respectively. While LSTMwHour receives only the hourly input dataset for a long duration (5040 hours) as input, LSTMwDpH receives the daily input dataset for 210 days together with the hourly input dataset for a short duration (210 hours) as input.

Table 3 Configuration of the layers of the CNN component of CNNpLSTM: The number of the channel of the feature map (NCH) or the number of the hidden nodes (NHN), the kernel size, the stride, and the padding width.

| Layer | NCH/NHN | Kernel | Stride | Padding |
|---|---|---|---|---|
| Conv1 | 8/16/32 | 6 | 3 | 3 |
| MaxPool1 | 8/16/32 | 4 | 2 | 2 |
| Conv2 | 16/32/64 | 4 | 2 | 2 |
| MaxPool2 | 8/16/32 | 4 | 2 | 2 |
| Conv3 | 32/64/128 | 4 | 4 | 2 |
| MaxPool3 | 32/64/128 | 4 | 2 | 2 |
| FC1 | 128 | - | - | - |

The CNN component of CNNsLSTM consists of three one-dimensional convolution layers. The configuration of the layers is given in Table 2. The configuration of each convolution layer of CNNsLSTM is set same as to that for 1D-CNN except that no pooling layer is utilized for CNNsSLTM because preliminary runs showed better accuracy without a pooling layer. The three NCHFs (8, 16, 32) are also used for CNNsLSTM. After each convolution layer, ReLU is applied as the activation function. Without flattening and a fully-connected layer, the feature map from the last convolution layer is directly given to the LSTM together with the hourly input dataset for a short duration (210 hours). Here, the size of the feature map from the last convolution layer is 210, which is the same as the input data length of LSTMwDpH. The LSTM component of CNNsLSTM also has a single layer and 30 of the hidden state.

The CNN component of CNNpLSTM has three convolution layers, three max-pooling layers, and one fully-connected layer. As shown in Table 3, the configuration of the convolution and max-pooling layers for CNNpLSTM is set to the same as that for 1D-CNN. Contrarily to CNNsLSTM, CNNpLSTM utilizes max-pooling layers because preliminary runs improve the results with the use of max-pooling. For CNNpLSTM, the LSTM component also has a single layer and 30 of the hidden state. The hourly input dataset for a long duration (5040 hours) and for a short duration (210 hours) are given to the CNN and LSTM components, respectively.

For all the architectures, the given dataset was separated into the training (2007-2015), validation (2016-2017), and test datasets (2018-2019). The learnable parameters are trained with the training dataset by means of the back propagation method with an optimization algorithm. Mean square error is utilized as the loss function to calculate the gradient. The optimization algorithm was set to Adaptive moment estimation (Adam: Kingma and Ba, 2014). From the training dataset, subsets, which are called mini-batches or batches, are generated by randomly extracting samples without duplication. The batch size was set to 512. An update of the learnable parameters is conducted with a batch. The number of the training is counted by an epoch, which is a set of updates with the entire training dataset. The overfitting of the model is examined with the validation dataset. Meanwhile, an early stopping method is utilized to avoid unnecessary training iterations. At each epoch, the model is run with the validation dataset after the training to calculate the loss for the validation dataset. The overfitting can be determined by the continuous increase in the loss for the validation dataset during a specific number of epochs. The number was set to be 30 in this study. The early stopping method stops the training when the overfitting is detected. Then, the model with the learnable parameters that minimize the loss for the validation data are utilized as the trained model.

The training process contains some randomness. The batches are generated by random sampling. In addition, the initial states of the learnable parameters are generally set by random values. Because such randomness affects the training results, this study performs the calibration process 100 times. Among the 100 trained mode, the best-trained model with respect to the loss for the validation period is utilized for verification with the test dataset. A deep learning framework for Python, Pytorch (Paszke *et al.*, 2019), was employed for the model implementation.

Finally, the model accuracy is evaluated by means of three statistical metrics: root mean square error (RMSE), correlation coefficient (R), and Nash–Sutcliffe efficiency (NSE). In addition, the model evaluation based on RMSE is also conducted for four ranges of river flow values, which were defined based on the hourly observed data: the low, middle, high, and peak flows. The low, middle and high flows indicate values below 25th percentile value, between 25th and 75th percentile values, and above 75th percentile value. Similarly, the peak river flow was defined as the flow above the 95th percentile value. The 25th, 75th, and 95th percentile values of flow are 257.8 $m^3$/s, 598.3 $m^3$/s, and 1293.4 $m^3$/s, respectively.

## 4. Results

The estimation accuracy by CNNsLSTM is compared to that by the four other architectures: 1D-CNN, LSTMwHour, CNNpLSTM, and LSTMwDpH. Figures 5-7 show correlation coefficient (R), NSE, and RMSE, respectively, obtained by each architecture for training, validation and test periods. Scatter plots between the observed river flow and the median values of the simulations are demonstrated in Figure 8 with corresponding R, NSE, and RMSE values. With respect to these statistic metrics, the new architectures, LSTMwDpH (Ishida et al., 2021) and CNNsLSTM (proposed this study), show better estimation accuracy compared to the other three conventional architectures, 1D-CNN, LSTMwHour, and CNNpLSTM. For instance, only LSTMwDpH and CNNsLSTM demonstrated the median correlation coefficient greater than 0.9, and the median NSE value greater than 0.8 for the test period in comparison to observed flows.

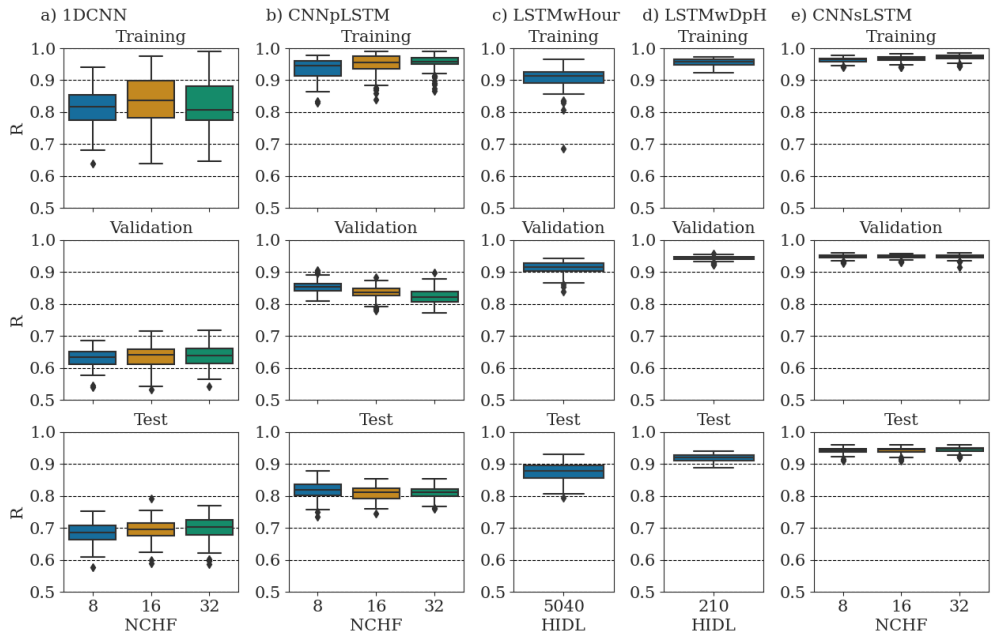

Figure 5. Correlation Coefficient values of deep learning modeling results by various approaches: 1D-CNN, CNNpLSTM, LSTMwHour, LSTMwDpH, and CNNsLSTM during the training, validation, and test periods. 1D-CNN, CNNpLSTM, and CNNsLSTM were run with different NCHF.

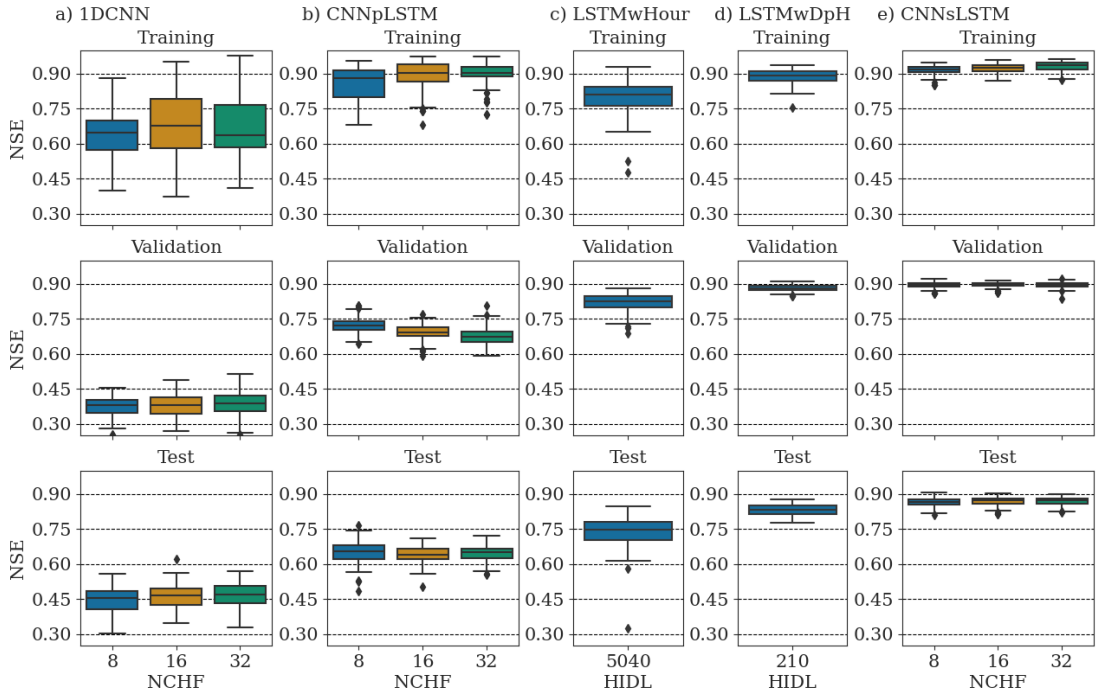

Figure 6. NSE values of deep learning modeling results by various approaches: 1D-CNN, CNNpLSTM, LSTMwHour, LSTMwDpH, and CNNsLSTM during the training, validation, and test periods. 1D-CNN, CNNpLSTM, and CNNsLSTM were run with different NCHF.

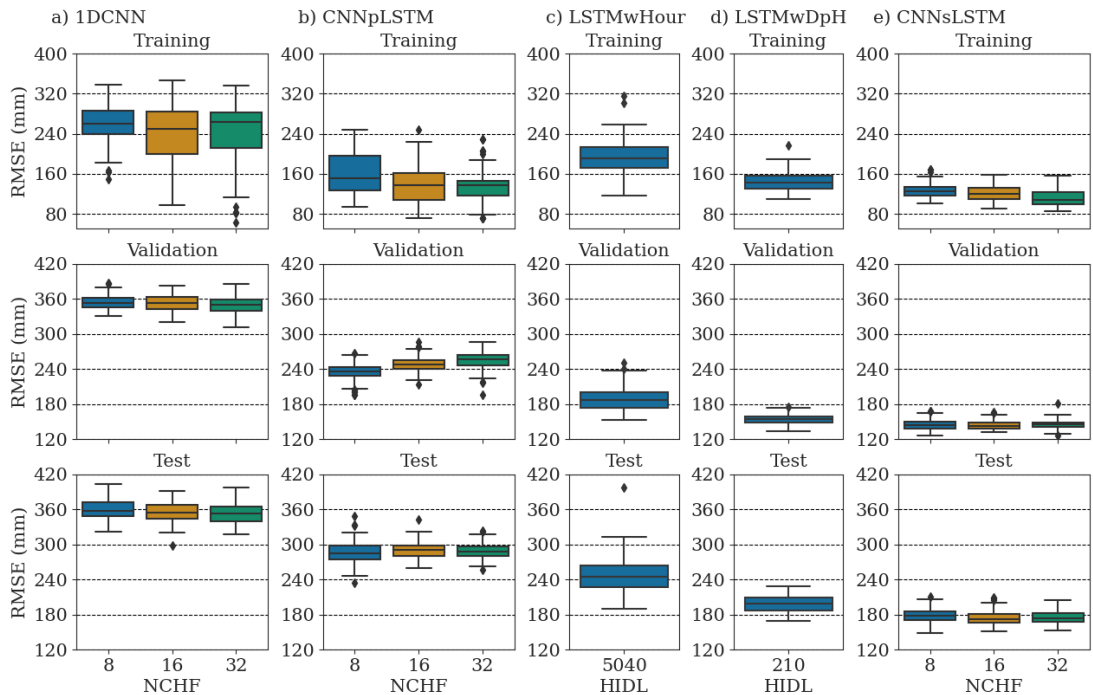

Figure 7. RMSE values of deep learning modeling results by various approaches: 1D-CNN, CNNpLSTM, LSTMwHour, LSTMwDpH, and CNNsLSTM during the training, validation, and test periods. 1D-CNN, CNNpLSTM, and CNNsLSTM were run with different NCHF.

Meanwhile, the proposed approach in this study, CNNsLSTM, shows relatively higher model accuracy compared to the LSTMwDpH. All the metrics obtained by CNNsLSTM are improved for the entire data during each of the three periods compared to those by LSTMwDpH, especially for the test period. The median values of the correlation coefficient, NSE, and RMSE for the test period are 0.943-0.946, 0.865-0.873, and 172.2-177.7 $m^3/s$, respectively, by CNNsLSTM with NCHF=8, 16, 32 while they are 0.920, 0.831, and 198.8 $m^3/s$, respectively, by LSTMwDpH. The improvements in the median correlation coefficient and NSE values are 0.023-0.027 and 0.034-0.042, respectively. The median value of RMSE for the test period is also reduced by 10.6%-13.4%.

The single use of a well-known architecture, 1D-CNN, shows the worst estimation accuracy for all the three periods among the five architectures. For example, the median of the NSE values obtained by 1D-CNN is smaller than 0.5 for the validation and test periods. Above 0.5 of NSE is considered acceptable according to Moriasi et al. (2007). 1D-CNN is not sufficient to be used for hourly-scale rainfall runoff modeling at the study watershed. CNNpLSTM shows better metrics for all the periods compared to 1D-CNN. During the training period, the model accuracy of CNNpLSTM with higher NCHF is comparable or even higher than that of LSTMwHour, LSTMwDpH, and CNNsLSTM. The medians of correlation coefficient and NSE by CNNpLSTM ranges from 0.946 to 0.959 and from 0.880 to 0.902, respectively, for the training period. However, CNNpLSTM clearly shows worse metrics for the validation and test periods compared to these three architectures. The medians of correlation coefficient by CNNpLSTM ranges from 0.823 to 0.854 for the validation period, and from 0.812 to 0.820 for the test period for NCHF=8-32. The medians of NSE by CNNpLSTM ranges from 0.692 to 0.721 for the validation period, and from 0.639 to 0.656 for the test period for NCHF=8-32. The trained results by CNNpLSTM seem to be more overfitted to the training dataset compared to the other architectures, especially with more NCHF. Then, the single use of another well-known architecture with the traditional approach of hourly input, LSTMwHour, shows the middle estimation accuracy among the five architectures. For example, the medians of correlation coefficient and NSE are 0.879 and 0.745, respectively, for the test period.

The model accuracy is also compared among the five architectures for the low, middle, high, and peak flows for the test period with respect to RMSE. The median values of RMSE are tabulated in Table 4. Same as the comparison with the entire flow data, LSTMwDpH and CNNsLSTM clearly show better model accuracy for each range of river flow compared to the other three architectures. Contrarily, the differences in the model accuracy between LSTMwDpH and CNNsLSTM changed among the ranges of river flow. The differences are relatively small for the low flow and the middle flow: The medians of RMSE by CNNsLSTM with NCHF=8-32 are 60.1 m$^3$/s, 60.3 m$^3$/s, and 58.9 m$^3$/s, respectively, for the low flow (< 25th %tile) while the median by LSTMwDpH is 63.5 m$^3$/s. The medians by CNNsLSTM with NCHF=8-32 are 95.3 m$^3$/s, 93.9 m$^3$/s, and 96.4 m$^3$/s, respectively for the middle flow (between 25the and 75th %tiles) while the median by LSTMwDpH is 100.0 m$^3$/s. CNNsLSTM improved RMSE by only 5.0%-7.2% for the low flow, and 3.6%-6.0% for the middle flow.

Table 4. Median values of RMSE ($m^3/s$) of deep learning modeling results by various approaches during the test period: 1D-CNN (NCHF=8, 16, 32), CNNpLSTM (NCHF=8, 16, 32), LSTMwHour, LSTMwDpH, and CNNsLSTM (NCHF=8, 16, 32) for the low, middle, high, and peak flow data.

|  | 1D-CNN | | | CNNpLSTM | | | LSTMwHour | LSTMwDpH | CNNsLSTM | | |
| --- | --- | --- | --- | --- | --- | --- | --- | --- | --- | --- | --- |
|  | 8 | 16 | 32 | 8 | 16 | 32 | - | - | 8 | 16 | 32 |
| Low flows | 174.2 | 179.3 | 160.8 | 129.4 | 142.9 | 138.8 | 83.9 | 63.5 | 60.1 | 60.3 | 58.9 |
| Middle flows | 216.9 | 219.6 | 211.1 | 177.5 | 187.3 | 184.6 | 118.3 | 100.0 | 95.3 | 93.9 | 96.4 |
| High flows | 596.5 | 589.5 | 594.4 | 476.8 | 482.6 | 474.0 | 442.1 | 354.1 | 311.5 | 302.0 | 303.3 |
| Peak flows | 1037.0 | 1029.1 | 1039.4 | 790.3 | 798.1 | 790.1 | 717.1 | 568.5 | 478.3 | 463.2 | 468.1 |

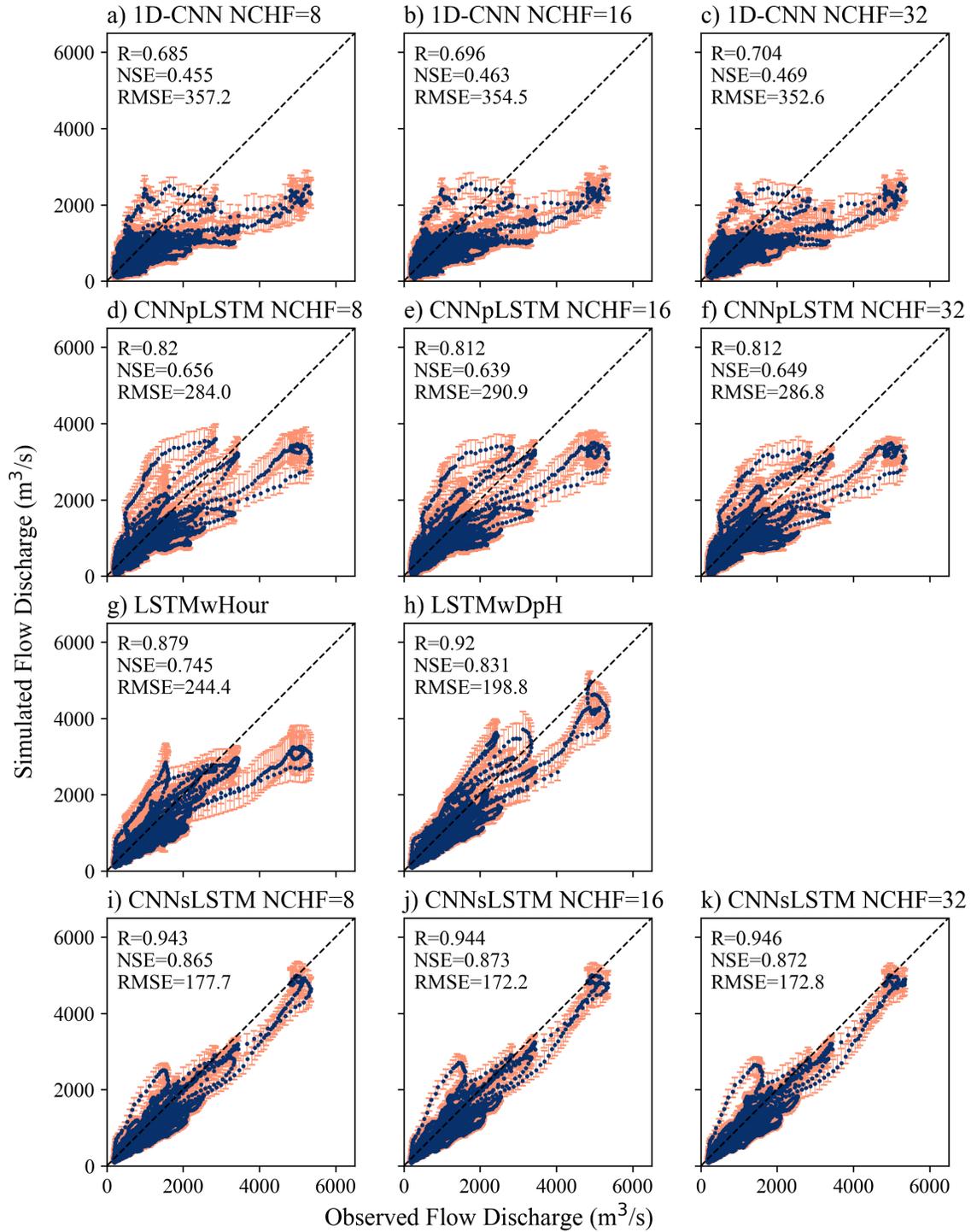

Figure 8. Scatter plots between the observed river flow and the median values of the simulations by various approaches (25th and 75th percentiles are shown by red color): 1D-CNN (NCHF=8, 16, 32), CNNpLSTM (NCHF=8, 16, 32), LSTMwHour, LSTMwDpH, and CNNsLSTM (NCHF=8, 16, 32).

Contrarily, relatively high improvements are found in RMSE for the high and peak flows. CNNsLSTM improved the model accuracy by more than 12% with respect to the median of RMSE compared to LSTMwDpH. For instance, the medians of RMSE by CNNsLSTM with NCHF=8-32 are 311.5 m$^3$/s, 302.0 m$^3$/s, and 303.3 m$^3$/s, respectively for the high flow ($\geqq$ 75th %tile) while the median by LSTMwDpH is 354.1 m$^3$/s. CNNsLSTM improved the median of RMSE by only 12.0%-14.7% for the high flow. In addition, higher improvements are found for the peak flow ($\geqq$ 95th %tile): The medians of RMSE by CNNsLSTM with NCHF=8-32 are 478.3 m$^3$/s, 463.2 m$^3$/s, and 468.1 m$^3$/s, respectively, while the median by LSTMwDpH is 568.5 m$^3$/s. CNNsLSTM improved the median of RMSE by 15.9%-18.5% for the peak flow.

## 5. Discussion

The proposed architecture, CNNsLSTM, showed clear improvements on the estimation accuracy of river flow compared to the single use of the well-known architectures, 1D-CNN and LSTMwHour, and the parallel architecture of CNN and LSTM, (CNNpLSTM). The clear improvements were found not only in the entire dataset, but also in different ranges of flow (classified as low, medium, high, and peak flows). CNNsLSTM also improved the estimation accuracy of river flow compared to the previously proposed architecture LSTMwDpH by Ishida et al. (2021). The improvements are relatively small for the low flow (< 25th %tile) and the middle flow (between 25the and 75th %tiles). However, the estimation of the high flow ($\geqq$ 75th %tile) and the peak flow ($\geqq$ 95th %tile) was clearly improved by CNNsLSTM. The median of RMSE for the test period was reduced by 12.0%-14.7% for the high flow, and by 15.9%-18.5% for the peak flow. Thus, CNNsLSTM would be valuable for flood management or hydraulic structure design, which require accurate high flow information.

In this study, IDL of LSTM was set to the same for CNNsLSTM and LSTMwDpH, which is 210. Both of them received 210 hours of hourly dataset as input. Then, LSTMwDpH received 210 days of daily data as input in addition to the hourly data. Contrarily, 5040 hours (210 days) of hourly data were given to the CNN component of CNNsLSTM. After the convolution operations with the activation function in the CNN component, the feature map with the length of 210 was obtained. Then the feature map was given to the LSTM component together with 210 hours of the data. This difference in input architecture resulted in major improvements on the estimation accuracy of the high and the peak flows by CNNsLSTM. Figure 9 shows the correlation between river flow and precipitation data at different times. Precipitation data within 10 hours mostly affected the flow in the study watershed. Although the hourly input data for a short duration are directly given to the LSTM component, above results indicate that the CNN component of CNNsLSTM also worked in learning the short-duration dependencies between the precipitation and flow data.

This study focused on hourly-scale rainfall-runoff modeling at a snow-dominated watershed, which requires the consideration of the long-duration dependencies between the meteorological input and the targeted river flow data. However, the proposed architecture CNNsLSTM may be useful for flood forecasting. Because CNNsLSTM shows better estimation accuracy especially for high and peak flows, it may be a better fit for flood forecasting studies, compared to the other four architectures.

The hourly data for a long duration were given to the CNN component as input in this study. Meanwhile, the hourly data for a short duration were given to the LSTM together with the feature map out from the CNN. The variables of the data for a long duration to the CNN component are the same as those for a short duration to the LSTM component. However, CNNsLSTM accepts different variables to be given to the CNN and LSTM components, respectively. For example, flood forecasting with LSTM sometimes utilizes river flow data together with precipitation data as input (e.g. Kao et al., 2020; Song et al., 2019; Xiang et al., 2020). When flow data are more frequently missing than precipitation data, it will be difficult to use flow data for a long duration as input to deep learning. In general, it may be more difficult to obtain a complete dataset for river flow data than meteorological data because nowadays gridded precipitation datasets are available, including atmospheric reanalysis datasets, and there are fewer missing parts in such datasets. In such a case, only the precipitation data can be utilized as input to the CNN component, and the precipitation data and flow data can be given together to the LSTM component for CNNsLSTM.

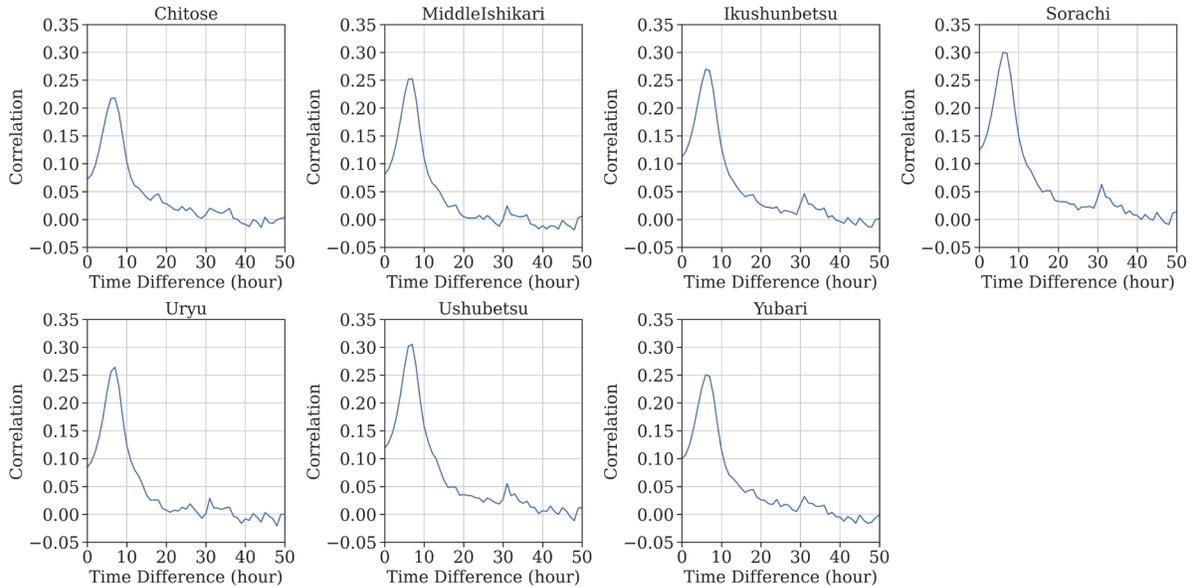

Figure 9. Correlation coefficient between the observed hourly river flow and the observed area-averaged precipitation at each region with time lags.

It is known that deep learning basically has an issue on extrapolation (Reichstein et al., 2019). In other words, it has difficulty to reliably estimate values that are above or below the training data. The final goal of this study is to develop a rainfall-runoff model to reliably estimate river flow at the hourly scale. The proposed architecture modeled river flow, especially for high and peak flows, with better accuracy. We believe that it can improve the reliability of rainfall-runoff modeling.

## 6. Conclusions

This study proposed an architecture consisting of a serial coupling of 1D-CNN and LSTM, which is called CNNsLSTM, for hourly-scale rainfall-runoff modeling at a snow-dominated watershed where there may be long-duration dependencies between meteorological and river flow datasets. In CNNsLTSM, the CNN component receives the hourly meteorological data for a long duration and the LSTM component receives the extracted features from 1D-CNN and the hourly meteorological data for a short-duration. As a case study, CNNsLSTM was implemented for hourly rainfall-runoff modeling at IRW. The meteorological data such as precipitation, air temperature, evapotranspiration, and long- and short-wave radiation were utilized as input. The flow data at an observation station near the river mouth were used as the target data. For evaluation of the estimation accuracy, CNNsLSTM was compared with three conventional architectures (1D-CNN, LSTMwHour, and CNNpLSTM) and the previously proposed architecture (LSTMwDpH) by Ishida et al. (2021). CNNsLSTM showed the clear improvements on the estimation accuracy compared to the three conventional architectures. In addition, the proposed model also improved the estimation accuracy of river flow compared to LSTMwDpH. Although the improvements were relatively small for the low flow (< 25th %tile) and the middle flow (between 25the and 75th %tiles), the estimation of the high flow ($\geqq$ 75th %tile) and the peak flow ($\geqq$ 95th %tile) was clearly improved. The median of RMSE for the test period was reduced by 12.0%-14.7% for the high flow, and by 15.9%-18.5% for the peak flow. The results indicate that the proposed architecture CNNsLSTM would be valuable for flood management and hydraulic structure design purposes mainly under climate change, which require accurate high river flow information river flow to be estimated from meteorological datasets.

### Acknowledgement

The radar raingauge-analyzed precipitation (RRAP) data were obtained from Japan Meteorological Business Support Center.

### References

Chen, R., Wang, X., Zhang, W., Zhu, X., Li, A., Yang, C., 2019. A hybrid CNN-LSTM model for typhoon formation forecasting. Geoinformatica 23, 375–396. https://doi.org/10.1007/s10707-019-00355-0

Copernicus Climate Change Service, 2017. ERA5: Fifth generation of ECMWF atmospheric reanalyses of the global climate . Copernicus Climate Change Service Climate Data Store (CDS) [WWW Document]. URL https://cds.climate.copernicus.eu/cdsapp#!/home (accessed 2.11.21).

Devia, G.K., Ganasri, B.P., Dwarakish, G.S., 2015. A Review on Hydrological Models. Aquatic Procedia 4, 1001–1007. https://doi.org/10.1016/j.aqpro.2015.02.126

G, Swapna, Kp, S., R, Vinayakumar, 2018. Automated detection of diabetes using CNN and CNN-LSTM network and heart rate signals. Procedia Comput. Sci. 132, 1253–1262. https://doi.org/10.1016/j.procs.2018.05.041

Gers, F.A., Schmidhuber, J., Cummins, F., 2000. Learning to forget: continual prediction with LSTM. Neural Comput. 12, 2451–2471. https://doi.org/10.1162/089976600300015015

Hersbach, H., Bell, B., Berrisford, P., Hirahara, S., Horányi, A., Muñoz‐Sabater, J., Nicolas, J., Peubey, C., Radu, R., Schepers, D., Simmons, A., Soci, C., Abdalla, S., Abellan, X., Balsamo, G., Bechtold, P., Biavati, G., Bidlot, J., Bonavita, M., Chiara, G., Dahlgren, P., Dee, D., Diamantakis, M., Dragani, R., Flemming, J., Forbes, R., Fuentes, M., Geer, A., Haimberger, L., Healy, S., Hogan, R.J., Hólm, E., Janisková, M., Keeley, S., Laloyaux, P., Lopez, P., Lupu, C., Radnoti, G., Rosnay, P., Rozum, I., Vamborg, F., Villaume, S., Thépaut, J., 2020. The ERA5 global reanalysis. Quart. J. Roy. Meteor. Soc. 146, 1999–2049. https://doi.org/10.1002/qj.3803

Hochreiter, S., Schmidhuber, J., 1997. Long short-term memory. Neural Comput. 9, 1735–1780. https://doi.org/10.1162/neco.1997.9.8.1735

Ishida, K., Kiyama, M., Ercan, A., Amagasaki, M., Tu, T., 2021. Multi-time-scale input approaches for hourly-scale rainfall–runoff modeling based on recurrent neural networks. J. hydroinformatics. https://doi.org/10.2166/hydro.2021.095

Kao, I.-F., Zhou, Y., Chang, L.-C., Chang, F.-J., 2020. Exploring a Long Short-Term Memory based Encoder-Decoder framework for multi-step-ahead flood forecasting. J. Hydrol. 583, 124631. https://doi.org/10.1016/j.jhydrol.2020.124631

Kim, T.-Y., Cho, S.-B., 2019. Predicting residential energy consumption using CNN-LSTM neural networks. Energy 182, 72–81. https://doi.org/10.1016/j.energy.2019.05.230

Kingma, D.P., Ba, J., 2014. Adam: A Method for Stochastic Optimization. arXiv [cs.LG].

Kratzert, F., Klotz, D., Brenner, C., Schulz, K., Herrnegger, M., 2018. Rainfall–runoff modelling using Long Short-Term Memory (LSTM) networks. Hydrol. Earth Syst. Sci. 22, 6005–6022. https://doi.org/10.5194/hess-22-6005-2018

Le, T., Vo, M.T., Vo, B., Hwang, E., Rho, S., Baik, S.W., 2019. Improving Electric Energy Consumption Prediction Using CNN and Bi-LSTM. NATO Adv. Sci. Inst. Ser. E Appl. Sci. 9, 4237. https://doi.org/10.3390/app9204237

Li, S., Xie, G., Ren, J., Guo, L., Yang, Y., Xu, X., 2020. Urban PM2.5 Concentration Prediction via Attention-Based CNN–LSTM. NATO Adv. Sci. Inst. Ser. E Appl. Sci. 10, 1953. https://doi.org/10.3390/app10061953

Li, W., Kiaghadi, A., Dawson, C., 2021. High temporal resolution rainfall–runoff modeling using long-short-term-memory (LSTM) networks. Neural Comput. Appl. 33, 1261–1278. https://doi.org/10.1007/s00521-020-05010-6

Li, X., Li, J., Qu, Y., He, D., 2019. Gear Pitting Fault Diagnosis Using Integrated CNN and GRU Network with Both Vibration and Acoustic Emission Signals. NATO Adv. Sci. Inst. Ser.

E Appl. Sci. 9, 768. https://doi.org/10.3390/app9040768

Liao, K., Zhao, Y., Gu, J., Zhang, Y., Zhong, Y., 2021. Sequential Convolutional Recurrent Neural Networks for Fast Automatic Modulation Classification. IEEE Access 9, 27182–27188. https://doi.org/10.1109/ACCESS.2021.3053427

Liu, M., Huang, Y., Li, Z., Tong, B., Liu, Z., Sun, M., Jiang, F., Zhang, H., 2020. The Applicability of LSTM-KNN Model for Real-Time Flood Forecasting in Different Climate Zones in China. Water 12, 440. https://doi.org/10.3390/w12020440

MLIT, 2021. . Water Information System. URL http://www1.river.go.jp/ (accessed 6.3.21).

Moriasi, D., Arnold, J., Van Liew, M.W., Bingner, R., Harmel, R.D., Veith, T.L., 2007. Model Evaluation Guidelines for Systematic Quantification of Accuracy in Watershed Simulations. Transactions of the ASABE 50, 885–900. https://doi.org/10.13031/2013.23153

Paszke, A., Gross, S., Massa, F., Lerer, A., Bradbury, J., Chanan, G., Killeen, T., Lin, Z., Gimelshein, N., Antiga, L., Desmaison, A., Kopf, A., Yang, E., DeVito, Z., Raison, M., Tejani, A., Chilamkurthy, S., Steiner, B., Fang, L., Bai, J., Chintala, S., 2019. PyTorch: An Imperative Style, High-Performance Deep Learning Library, in: Wallach, H., Larochelle, H., Beygelzimer, A., d\textquotesingle Alché-Buc, F., Fox, E., Garnett, R. (Eds.), Advances in Neural Information Processing Systems 32. Curran Associates, Inc., pp. 8024–8035.

Reichstein, M., Camps-Valls, G., Stevens, B., Jung, M., Denzler, J., Carvalhais, N., Prabhat, 2019. Deep learning and process understanding for data-driven Earth system science. Nature 566, 195–204. https://doi.org/10.1038/s41586-019-0912-1

Shen, C., 2018. A transdisciplinary review of deep learning research and its relevance for water resources scientists. Water Resour. Res. 54, 8558–8593. https://doi.org/10.1029/2018wr022643

Shi, X., Chen, Z., Wang, H., Yeung, D.-Y., Wong, W.-K., Woo, W.-C., 2015. Convolutional LSTM Network: A Machine Learning Approach for Precipitation Nowcasting. arXiv [cs.CV].

Song, T., Ding, W., Wu, J., Liu, H., Zhou, H., Chu, J., 2019. Flash Flood Forecasting Based on Long Short-Term Memory Networks. Water 12, 109. https://doi.org/10.3390/w12010109

Sun, J., Di, L., Sun, Z., Shen, Y., Lai, Z., 2019. County-Level Soybean Yield Prediction Using Deep CNN-LSTM Model. Sensors 19. https://doi.org/10.3390/s19204363

Tian, Y., Xu, Y.-P., Yang, Z., Wang, G., Zhu, Q., 2018. Integration of a Parsimonious Hydrological Model with Recurrent Neural Networks for Improved Streamflow Forecasting. Water 10, 1655. https://doi.org/10.3390/w10111655

Van, S.P., Le, H.M., Thanh, D.V., Dang, T.D., Loc, H.H., Anh, D.T., 2020. Deep learning convolutional neural network in rainfall--runoff modelling. Journal of Hydroinformatics 22, 541–561.

Vidal, A., Kristjanpoller, W., 2020. Gold volatility prediction using a CNN-LSTM approach. Expert Syst. Appl. 157, 113481. https://doi.org/10.1016/j.eswa.2020.113481

Xiang, Z., Yan, J., Demir, I., 2020. A rainfall‐runoff model with LSTM‐based sequence‐to‐sequence learning. Water Resour. Res. 56. https://doi.org/10.1029/2019wr025326

Xu, M., Yin, Z., Wu, M., Wu, Z., Zhao, Y., Gao, Z., 2020. Spectrum Sensing Based on Parallel CNN-LSTM Network, in: 2020 IEEE 91st Vehicular Technology Conference (VTC2020-Spring). ieeexplore.ieee.org, pp. 1–5. https://doi.org/10.1109/VTC2020-Spring48590.2020.9129229

Xu, T., Liang, F., 2021. Machine learning for hydrologic sciences: An introductory overview. WIREs Water. https://doi.org/10.1002/wat2.1533

Yang, R., Singh, S.K., Tavakkoli, M., Amiri, N., Yang, Y., Karami, M.A., Rai, R., 2020. CNN-LSTM deep learning architecture for computer vision-based modal frequency detection. Mech. Syst. Signal Process. 144, 106885. https://doi.org/10.1016/j.ymssp.2020.106885

Yu, J., Zhang, X., Xu, L., Dong, J., Zhangzhong, L., 2021. A hybrid CNN-GRU model for predicting soil moisture in maize root zone. Agric. Water Manage. 245, 106649. https://doi.org/10.1016/j.agwat.2020.106649

Zang, H., Liu, L., Sun, L., Cheng, L., Wei, Z., Sun, G., 2020. Short-term global horizontal irradiance forecasting based on a hybrid CNN-LSTM model with spatiotemporal correlations. Renewable Energy 160, 26–41. https://doi.org/10.1016/j.renene.2020.05.150